\documentclass[twocolumn]{article}

\usepackage[width=17cm,height=22cm]{geometry}
\usepackage[english]{babel}
\usepackage[utf8]{inputenc}
\usepackage{fancyvrb}
\usepackage{authblk,graphicx}
\usepackage{hyperref}
\usepackage{color}
\usepackage{eurosym}
\usepackage{booktabs}

\usepackage{tikz}
\usetikzlibrary{shapes,arrows}
\usepackage{amsmath,bm,times}

\title{A multi-series framework for demand forecasts in E-commerce}
\author[1]{R\'emy Garnier }
\author[2]{Arnaud Belletoile}
\affil[1]{Universit\'e Paris Seine, Laboratoire AGM UMR 8088,
95000 Cergy-Pontoise \& CDiscount 33000 Bordeaux, France.}
\affil[2]{CDiscount 33000 Bordeaux, France}

\begin{document}
\maketitle

\begin{abstract}
  Sales forecasts are crucial for the E-commerce business. State-of-the-art techniques typically apply only univariate methods to make prediction for each series independently. However, due to the short nature of sales times series in E-commerce, univariate methods don't apply well. In this article, we propose a global model which outperforms state-of-the-art models on real dataset. It is achieved by using Tree Boosting Methods that exploit non-linearity and cross-series information. We also proposed a preprocessing framework to overcome the inherent difficulties in the E-commerce data. In particular, we use different schemes to limit the impact of the volatility of the data.  
\end{abstract}

\medskip

\noindent\textbf{Keyword}: E-Commerce, Demand Forecasts, Boosting, Applied Machine Learning.

\section{Introduction}
\label{sec: Intro}

 The recent rise of E-commerce has created a need for operational product-level demand forecasts. Indeed, modern standards in logistics require \emph{just-on time} resupply. Better accuracy on forecasts can lead to huge savings and cost reductions.

However, the business environment in E-commerce makes this prediction complex because of the volatility of the sales. For instance, sales are affected by holiday effects, competitor behaviours, pricing changes,... Demand data carry various challenges such as non-stationary historical data, short times series, cannibalisation effects.

There are natural groupings of products, where items of the same type, sub type, mark or  price segment fall into the same group. In those group, the key properties of each products are close to each other. For instance, the product of the category 'Toys' will share the same seasonal behavior, and theirs sales will increase at Christmas.

The existing methods generally treat different series separately. It may work well with the physical retail, but the rapid rotation of products and the volatility of the demand in online retail create a need to provide models which share information between times series.  \cite{yelland2010bayesian,chapados2014effective,trapero2015identification,bandara2019sales}.

In this study, we will propose a framework for real-world demand forecasting problem in E-commerce. Our goal is to exploit the correlation between series to improve the accuracy of predictions. In particular, we want to tackle the problem of short history of time series.

In Section \ref{sec:Contexte}, we define formally the problem and propose a rapid review of previous work on the field. We present a preprocessing of the data in section \ref{sec:preprocessing}. In section \ref{sec: model}, we present the boosting model which gives us the best performance. Finally, we present the setup and results of our experiment on a real-world data-set on section \ref{sec:experiment}.

\section{Context}
\label{sec:Contexte}
\subsection{Formulation}

We have a set $I$ of products, divided in K different categories $I_k$ such that $I = \biguplus_{k} I_k$ ($I$ disjoint union of $I_k$).
We also have $N$ count times series $(y_{i,t})$, where $y_{i,t}$ represent the number of sales of the product $i \in I$ during the week $t$. This series are observed during $T$ weeks. 

The support of this series, i.e. the number of non-zero week for each series is relatively small compared to $T$. This means that we don't observe a lot of history for each product individually. We suppose that the series follow a seasonality of period $\tau$, mostly annual ($\tau = 52$), although an entire period $\tau$ is seldom observed.

Some externals features $Z=((z_{i,t})_i)_t$ are important. Three types of covariates may be used:

\begin{itemize}
    \item \textbf{Temporal features:} Covariates that depend on the date $t$ only. They are common for all  products. For instance, special events (Christmas, Black Friday) and the weather-related covariates  fall into this category.
    \item  \textbf{Longitudinal features: }Covariates that depend on the product $i$ only. For instance the type of product, its mark. Longitudinal features allow to produce a hierarchy of products.
    \item \textbf{Mixed features}: Covariates that depend on both. For instance, prices of a product may vary every week. 
\end{itemize}

Our objective is to forecast  values of this series for an horizon $h$. More formally, we wish to develop a prediction model $f$, such that, if we consider the past sales of a product i $y_{i,:t} = (y_{i,0}, \dots, y_{i,t})$, the value $f(y_{i,:t}, z_{i,t}, \theta)$ is an estimator of $y_{i,t+h}$ for a set of learnable parameters $\theta$.

\subsection{Related Work}

A large amount of work has been published concerning the demand forecasting methods, for different applications (facilities, physical and online retail,...). The most widely used methods are  classical times series models such as ARIMA models \cite{ediger2007arima}and exponential smoothing variants \cite{taylor2003}. However forecasting in the E-commerce space commonly needs to address challenges such as irregular sale trends, presence of highly bursty and sparse sale data,  \textit{etc}. Some of those limitations can be overcome through modified likelihood function and extended linear models \cite{seeger2016bayesian}. But this methods fails to achieve good performance when the series are small.

Other regression methods have been used, such as generalized additive methods \cite{pierrot2011short}, support vector machines \cite{chen2004load} and Recurrent Network \cite{borovykh2017conditional}. All this method performs only univariate forecasting and therefore run into the same problems.

Recently, neural networks \cite{bandara2019sales} have been proposed to use cross-series information for the specific purpose of E-commerce. They adapt a Long Short-Term Memory Neural Network (LSTM) architecture to treat all the series at the same time. They also separate effects of longitudinal and temporal features and seem to have a good performance. This suggests that  non-linearity is important for modelling such data.

 Bayesian hierarchical models are another promising models \cite{yelland2010bayesian,chapados2014effective}. This framework fits a simple model for each time series with some constraints on the learnable parameters for the models. This constraints are based on prior assumption on the distribution of the learnable parameters among the different products . For instance, we can impose a prior distribution of the effect of some covariates. This allows to share information between times series and to separate the effects of each covariates. Moreover, it gives confidence bounds on our prediction, 

\section{Data Preprocessing}
\label{sec:preprocessing}
\subsection{Preprocessing of sales features}
\label{sec:smooth}
There are two types of issues with sales data in E-commerce. The first one is the presence of abnormally low values, or 'fake zeros'. Those low values can be due to stock shortages, network issues or modification of the search engine on the website. As we want to predict the demand, we have to identify and replace this values. 'Fake zeros' can be identified through different methods, using stock information and different threshold. We replace them using a standard univariate prediction algorithm based on classical times series methods on each series.

The second issue is the presence of abnormaly high values. These values are informative, because they provide us with information on the effect of sales. However, those values are problematic when used as lags features, because they may suggest a higher level of sales than expected, or misinform about trend and seasonality. Therefore, we construct 'smoothed sales' $x_{i,t}$ eliminating the values which exceed $\gamma$ times the standard variation. More precisely: 

\begin{itemize}
    \item We calculate for each product $i$ a moving average series, and a moving standard deviation
    \[ \overline{y_{i,t}} = \frac{1}{M} \sum_{k=0}^M y_{i,t-k}\]  \[\overline{\sigma_{i,t}} = \left(\frac{1}{M} \sum_{k=0}^M (y_{i,t-k} - \overline{y_{i,t}} )^2\right)^{\frac{1}{2}}\].
    \item If $y_{i,t}  > \overline{y_{i,t}} + \gamma \overline{\sigma_{i,t}}$, then $x_{i,t} = \overline{y_{i,t}} + \gamma \overline{\sigma_{i,t}}$. 
    \item Otherwise $x_{i,t} = y_{i,t}$.
\end{itemize}

\subsection{Trend and seasonality}
\label{sec: seas}
We want to enrich the features with information about trend and seasonality. The goal is to produce features which can be compared between products. This will allow to use them as a global features for all products. 

First, we generate some normalized  trend features using the smoothed values. We use both an annual trend calculated by regression over the previous year data (when available) and a local trend.

The treatment of seasonality is more complex, because of the short nature of this series. We use a variant of the procedure describe in \cite{kumar2002clustering} to produce a seasonality factor for each product. Let us sketch this procedure.

First, we normalize the sales numbers for each year. We want to ensure that each products has the same mean level. For each product $i$, considering $N_i$ the number of weeks during the year during which the product was actually on sales, we note for a date $t$ in this year (i.e $t \in {0, \dots, \tau -1 }$):  
\[ x^{std}_{t,i} =  \frac{N_i}{\tau}\cdot \frac{x_{t,i} }{\displaystyle \sum_{t=0}^{\tau}x_{t,i}}\]
Second, we compute the mean of the standardize values in each categories $I_k$. We therefore have $K$ standardized seasonality for each product categories. The core idea is to suppose that there is a common multiplicative seasonality $s_{I_k}(t)$ for all product of this categories. Therefore, if the date on which the product was placed into the market are uniformly distributed, the calculated mean is directly proportional to the seasonality. 

However, due to the erratic natures of E-commerce sales data, at this step, the calculated seasonality are often not informative enough and contains some noise.
That is why we use a time series clustering algorithm to cluster the seasonality of the different categories. This clustering is based on the Euclidian distance between seasonality patterns,but also take into account the variance of the seasonality in the category.

 We finally produce a small numbers of seasonality patterns. This patterns allows us to produce a seasonality feature for each product according to its category, even if we don't have any information on its previous sales.

\subsection{Others features treatment}

\paragraph{Encoding of categorical features}
Longitudinal features are often categorical features, so we need to encode them to use them with numerical algorithm. However, due to the high number of categories, standard One-hot-encoding creates a lot of features and imposed to use very simple models for the regression.

Two possibilities sill remains for more complex models. First, the use of an ordinal encoding is simple and easy to implement, but it introduce an order on features, which doesn't really make sense.

Second, it is also possible to hash the features in order to obtain a small number of columns. This avoid partially the ordering of the features. However, it now becomes harder to make the importance of each value explicit.

\paragraph{Unpredictible features}

Some mixed or temporal features, like weather or prices cannot be for prediction, because they cannot be predicted for the horizon where we want to predict information. 
However, this features can be used to train the model on the past data, in order to explain abnormaly low (or high) values in the past. We can then performs prediction using a guess on the future values. For instance, we can take the seasonal value of weather features, or the mean observed price of the past data. This scheme has a weakness : the fact that we use exact past values leads machine learning algorithm to give to much importance to this features.

\section{Model}
\label{sec: model}
\subsection{Learning schemes}

We consider our problem of multiple time series forecasting as a regression problem. Our objective is the sale values corrected from 'fake zeros' at the horizon $y_{i,t+h}$. We use the past values of the smoothed sales as features, as described in \ref{sec:smooth}. We therefore have a prediction

\[ 
\widehat{y_{i,t+h}} =  f(x_{i,:t}, z_{t + h,i}, \theta) \]

The hypothesis is that $x_{i,t}$ represent 'normal' level of sales. It is suppose to remove the effects of punctual effects, like special offers. Features $z_{i,t }$ gives us information about the difference  $\delta_{i,t}=  y_{i,t} - x_{i,t}$. Therefore, we prefer using smoothed lagged values $x_{i,t}$ as features instead of lagged $y_{i,t}$.

However, we cannot completely separate the estimation of $\delta_{i,t}$ and $x_{i,t}$, because the value of $\delta_{i,t}$ strongly depend on the level of $x_{i,t}$. And this is hard to distinguish features that affects only $\delta_{i,t}$ from the features which affect $x_{i,t}$.

We used as learning set the values of the tuples $(x_{i,:t}, z_{t + h,i})$ for all products $i$ before a given date. Hyper-parameters are selected using a simple validation period.  

\begin{figure*}[ht]
    \centering
\pgfdeclarelayer{background}
\pgfdeclarelayer{foreground}
\pgfsetlayers{background,main,foreground}

\tikzstyle{sensor}=[draw, fill=blue!20, text width=7em, 
    text centered, minimum height=4em]
\tikzstyle{ann} = [above, text width=5em]
\tikzstyle{naveqs} = [sensor, text width=6em, fill=red!20, 
    minimum height=16em, rounded corners]
\def\blockdist{3.4}
\def\edgedist{2.5}

\begin{tikzpicture}
    \node (naveq) [naveqs] {Machine Learning function $f$};
  
    \path (naveq.-145)+(-2.8*\blockdist,0) node (HSD) [sensor] {Historical Sales Data};
    \path (naveq.-145)+(-1.8*\blockdist,0) node (NZSD) [sensor] {'Fake-zeros' free sales data $y_{i,t}$ };
    \path (naveq.-145)+(-0.8*\blockdist,0) node (SSD) [sensor] {Smoothed sales data $x_{i,t}$};
    \path (naveq.-115)+(-0.8*\blockdist,0) node (ST) [sensor] {Trend and Seasonality};
    \path [draw, ->] (HSD) -- node [above] {} (NZSD) ; 
    \path [draw, ->] (NZSD) -- node [above] {} (SSD) ; 
    \path [draw, ->] (SSD) -- node [above] {} (naveq.west |- SSD) ; 
    \path [draw, ->] (SSD) -- node [above] {} (ST) ; 
    \path [draw, ->] (ST) -- node [above] {} (naveq.west |- ST) ; 
    
    \path (naveq.145)+(-2.8*\blockdist,0) node (accel) [sensor] {Categorical Features};
    \path (naveq.145)+(-0.8*\blockdist,0) node (ECF) [sensor] {Encoded Categorical Feat.};
    \path [draw, ->] (accel) -- node [above] {Encoding} (ECF) ; 
    \path [draw, ->] (ECF) -- node [above] {} (naveq.west |- ECF) ; 

    \path (naveq.115)+(-2.8*\blockdist,0) node (gyros) [sensor] {Numerical Features};
    \path (naveq.115)+(-0.8*\blockdist,0) node (CF) [sensor] {Completed Numerical Features};
    \path [draw, ->] (gyros) -- node [above] {Missing Data Inputation} (CF) ; 
    \path [draw, ->] (CF) -- node [above] {} (naveq.west |- CF) ; 

    \path (naveq.east)+(0.5*\blockdist,0) node (c)  {Output};
    \path [draw,->] (naveq.east) -- node [above] {} (c); 
    
    \path (naveq.south) +(0,-0.5*\blockdist) node (d)  {};
    \path [draw] (naveq.south) -- node [above] {} (d.center);
    \path (d) +(-2.14*\blockdist,0) node (e)  {};    
    \path [draw] (d.center) -- node [above] {Objective (Poisson Loss Minimisation)} (e.center);
    \path [draw,->] (e.center) -- node [above] {} (NZSD.south);
    \path (HSD)+(0,-1.3) node (INS) {Input};
    
    \begin{pgfonlayer}{background}
        \path (gyros.west |- naveq.north)+(-0.5,0.3) node (a) {};
        \path (INS.south -| naveq.east)+(+0.3,-0.2) node (b) {};
        \path (gyros.north west)+(-0.2,0.2) node (a) {};
        \path (HSD.south -| gyros.east)+(+0.2,-0.2) node (b) {};
        \path[fill=blue!10,rounded corners, draw=black!50, dashed]
            (a) rectangle (b);
    \end{pgfonlayer}
\end{tikzpicture}

    \caption{General prediction Scheme}
    \label{fig:Scheme}
\end{figure*}
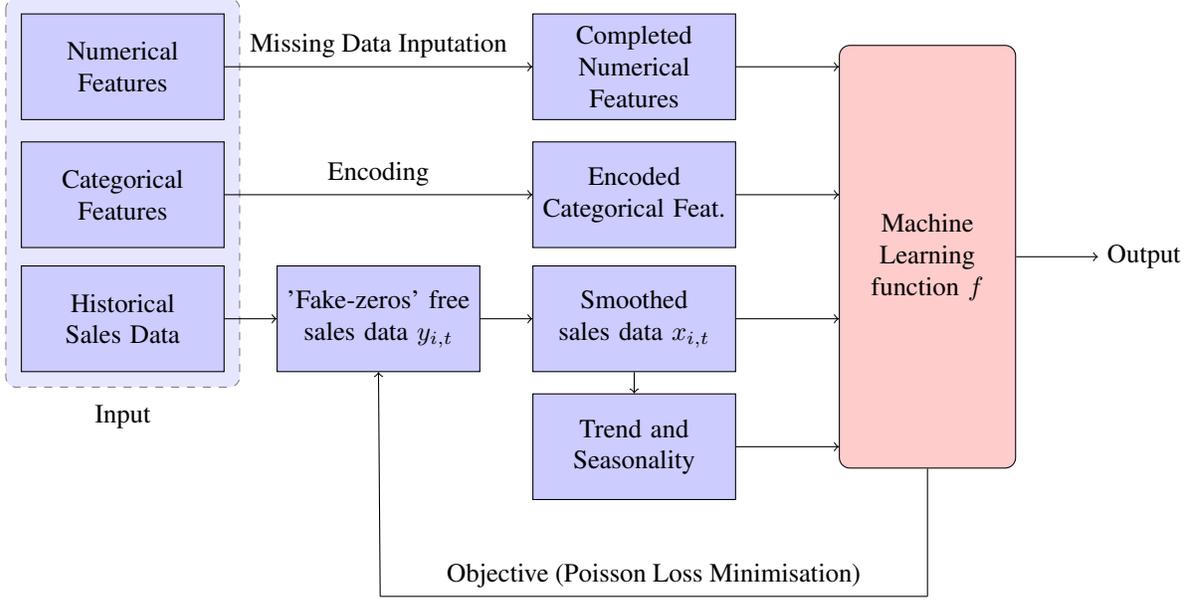

We sum up everything on the figure \ref{fig:Scheme}.

\subsection{Loss}

We distinguish the evaluation metrics, used for the evaluation of the our prediction, and the learning metrics, used in our models to evaluate the dispersion of the times series.

We used the standard Rooted Mean Squared Error (RMSE) and Mean Absolute Error (MAE) as evaluation metrics, using the price of the product $p_i$ as weight.

\[ RMSE(f,\theta) = \sqrt[]{\frac{1}{n}\sum_{i \in I} p_i^2 \cdot (y_{i,t+h}- f(x_{i,:t}, z_{t + h,i}, \theta))^2 }\]

\[ MAE(f,\theta) = \frac{\sum_{i \in I} p_i \cdot | y_{i,t+h}- f(x_{i,:t}, z_{t + h,i}, \theta)|}{\sum_{i \in I}  p_i \cdot f(x_{i,:t}, z_{t + h,i}, \theta) } \]

This metrics are commonly used in supply chain forecasting. However, RMSE are sensitive to extreme values, and both metrics tends to underestimate prediction.

Therefore, we proposed to use a Poisson Loss as learning metrics. This metrics has already been used in \cite{borovykh2017conditional} . We suppose that $y_{i,t}$ follows a Poisson distribution of parameter $$\widehat{y_{i,t}}=f(x_{i,:t}, z_{t ,i}, \theta)$$  The criterion we want to optimize is then the log-likelihood of the value $y_{i,t}$, or Poisson-loss:

\[ \mbox{Poisson}(f,\theta) = \sum_{i \in I} \widehat{y_{i,t+h}} - y_{i,t} \log(\widehat{y_{i,t+h}}) \]

It is a natural choice for three reasons. 

First, we observed that the sales time series are strongly heteroscedastic, and that the local variance of the series is strongly correlated with the local mean of the time series.\\
Second, it allows us to limit the effects of the presence of outliers in our data. Indeed, higher values are more likely than in a gaussian white noise modelling for instance.\\
Third, the positive integer values are naturally modelled by counting process. We can suppose, that for each week $t$ and each product $i$, the client arrived following a Poisson process, and that the parameter of this Process change each week.\\

 \cite{borovykh2017conditional}.
\subsection{Algorithms}

On the one hand, the choice of the machine learning algorithm to compute $f$ and $\theta$ is crucial. On the one hand, it must be flexible enough to use different kind of features, and to select the more useful features. In particular, it should be able to resist to redundant or correlated features. On the other hand, it must be consistent enough to avoid over-fitting. Finally, due to the large number of series and features, it must be fast enough to handle large data.

We test different models. For each, we try to perform variable selection through simple validation. We also try to normalize the features in the different case.

\paragraph{Linear models}

It is possible to use different linear models with One-Hot Encoded categorical data. With Lasso or Elastic Net penalization, it performs a good variable selection. However, it doesn't model threshold and other non-linear effects. And it doesn't use cross-features effects.

\paragraph{Generalized Additive Models}
A standard generalisation of the Linear models are the Generalized additive models (GAM), often used in time series prediction \cite{hastie2017generalized}. It consists in the regression of a function on a spline base, which consist on simple function of the parameters. It allows to treat non-linear effects, but cross-features effects have to be imposed \emph{manually}. 
Here, we haven't been able to find a configuration of GAM models which offers good performances.

\paragraph{Random Forests Regression}

Random Forests are a type of bagging algorithms, which consists in the construction of different regression tree by boostrap, and then produce a prediction based on the predictions of the different trees. It allows to take threshold and cross-features effects into account. And it can be parallelized, which allows for a fast computation.

Random forests are well suited for the estimation of $f$, and therefore obtains good performance on the datasets.

\paragraph{Boosting Tree Algorithm}

Contrary to tree bagging methods, tree boosting methods implements a sequential pooling of the prediction of different trees. They have recently receive a lot of attention, due to their performance on real case. Here we mostly use XgBoost \cite{chen2016xgboost}, which is a fast gradient boosting implementation . 

Here,it keeps the advantages of Random Forests, but have better performance.  The price is a generally higher training time, because the training cannot be parallelized. XgBoost hyper-parameters are selected via validation. The scope of validation are presented on the table \ref{tab:tabcol}. We use early stopping to reduce the training time. 

\begin{table}[htbp]
  \centering
  \begin{tabular}{lcc}
    \toprule
    \textbf{Parameter} & \textbf{Min value} & \textbf{Max value}  \\ 
    \midrule
    learning rate & 0.01 & 0.3 \\ 
    min split loss & 0.01 & 0.2 \\ 
    max depth & 5 & 8\\ 
    round evaluation & 1000 & 5000 \\
    \bottomrule
  \end{tabular}
  \caption{XgBoost hyper-parameters range}
  \label{tab:tabcol}
\end{table}

We also tried to use LightGBM \cite{ke2017lightgbm}, which runs faster, but obtains slightly worst performances.

\section{Experiments}
\label{sec:experiment}

\subsection{Dataset}

\begin{figure}[htbp]
  \centering
  \includegraphics[width=8cm,height=5cm]{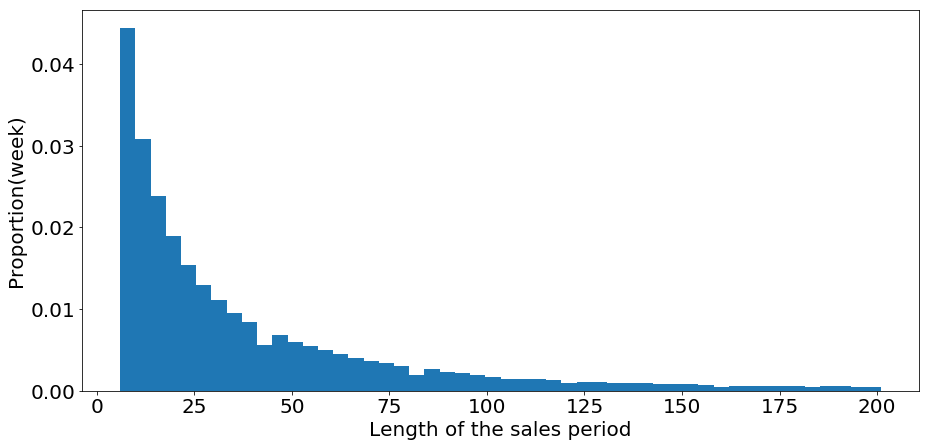}
  \caption{Repartition of sales period length for the product}
  \label{fig:code}
\end{figure}

We use our forecasting framework on a dataset collected from \emph{Cdiscount.com}. It collects the sales of $99305$ products, in $1031$ categories during approximately 4 years. On the figure \ref{fig:code}, we represent the repartition of the length of the sales for the products of the data-sets. A large proportion of them are sold during a short period.

We define our forecasting horizon $h$ as 6, and we train the model on the first 170 weeks, then use the 10 next weeks for validation of the hyper-parameters. To evaluate the model, we use the last 19 weeks. This weeks correspond to the lasts week of the year 2018 and the beginning of 2018. They contains therefore a lot of variability (Black Friday, Christmas, Winter Sales)

There are 3 sets of products, named A, B and C. The first ones regroups the products which sales the most, the last one the products which sales the least.

\subsection{Benchmark and XgBoost variants}

We compare our algorithms to both homemade and state-of-the-art benchmark. First, we use a state-of-the-art algorithm of the industry, called Benchmark. This algorithm performs a prediction for each series using a classification of the times series and business knowledge.

We also use a simple exponential smoothing algorithm as benchmark(ES) .

We also present the performance when we use different machine learning algorithm, than XgBoost, for instance Random Forests (RF).

Another advantage of the global method of prediction is that it allows \emph{cold-start} prediction, i.e. prediction on new series without history. In order to have fair evaluation with the benchmark, we remove the first 6 weeks of life of the products, where our algorithm are able to make prediction, but not benchmark algorithms. 

\subsection{Results}

Table \ref{sec: model} shows the performance of the prediction for two evaluation metrics for the total set of products, and for the different set A,B and C. The RMSE values are expressed in terms of k\euro{}. We present different version of our algorithm, depending on the encoding of categorical features (ordinal or hashing) and the use of the seasonality features (described in \ref{sec: seas}).

We can see that XgBoost outperforms the Benchmark for all categories. It reduces the MAE by approximately 5 \% of and the RMSE by 10 \%  on the whole data-sets.Globally, the relative gain is more important in RMSE, than in MAE, which shows that it mostly reduces the biggest gap of performances than it improves the average prediction. 

Ordinal Encoding, strangely, seems better than Hashing. And models with seasonality are better on the most sold product of the group A, which present the highest business impact.

\begin{table*}[!ht]
    \centering
    \begin{tabular}{lllc@{\hskip.8mm}c@{\hskip.12mm}@{\hskip.9mm}c@{\hskip.12mm}c@{\hskip.9mm}@{\hskip.12mm}c@{\hskip.9mm}c@{\hskip.12mm}@{\hskip.9mm}c@{\hskip.12mm}c}
        \toprule
        \multicolumn{3}{c}{\textbf{Framework}} &  \multicolumn{2}{c}{\textbf{All}} & \multicolumn{2}{c}{\textbf{A}} & \multicolumn{2}{c}{\textbf{B}} & \multicolumn{2}{c}{\textbf{C}}\\
        \midrule
        ML Algo. & Configuration & Encoding & RMSE   &  MAE  & RMSE   & MAE &  RMSE   & MAE &  RMSE   & MAE \\
        \midrule
        ES &   & & 3.83 & 1.09 & 5.68 & 1.03 & 1.12 & 1.06 & 1.28  & 1.31 \\
        RF & with seas. & Ordinal  & 3.09  & 0.831 & 5.27 & 0.796 & 1.66 & 0.892  & 1.32  & 0.92 \\
        XgBoost &  Poisson/with seas. & Ordinal& \textbf{2.67} &  \textbf{0.725} & \textbf{4.59} & \textbf{0.674}  & 1.41  & 0.801 & \textbf{1.20} & 0.874  \\
        XgBoost &  Poisson/without seas. & Ordinal & 2.76  & 0.730  & 4.72 & 0.681 & \textbf{1.39} &  \textbf{0.800} &  1.21 & \textbf{0.872} \\
        XgBoost &  Poisson/with seas. & Hashing &  2.78  & 0.728   & 4.75 & 0.685  & 1.41 & 0.816 & 1.21  & 0.893 \\
        XgBoost &  Poisson/without seas. & Hashing  & 2.79 & 0.740 & 4.77  & 0.689  &  1.40 & 0.817  & 1.22  & 0.893 \\
        Benchmark &  &  &  3.01 & 0.758 & 4.97 & 0.688  & 1.77 & 0.907 &  1.56 & 0.982 \\
        \bottomrule
    \end{tabular}
    \label{tab:result}
    \caption{Comparison of different models.}
\end{table*}

If we look closely at the performance, we see that our algorithm is particularly performing during the first week of the product cycle. We present this results in the $\ref{tab:start}$ for the Benchmark and our framework with the seasonality and ordered features. The high variability of RMSE is due to the small number of product concerned and the high variability of the studied period. Nevertheless, we can observe that, at the beginning of the product cycle, our framework strongly outperforms the benchmark. We decreases for instance the MAPE by $24,0$\% and the RMSE by $42,5$ \% for the product with 10 weeks of historical data.
This difference decrease with time, as the benchmark gain sufficient history for its prediction.

\begin{table}[!htbp]
    \centering
    \begin{tabular}{ccccc}
        \toprule
        \textbf{Product cycle} &  \multicolumn{2}{c}{\textbf{Framework}} & \multicolumn{2}{c}{\textbf{Benchmark}} \\
        \midrule
            Length & RMSE & MAE & RMSE & MAE \\
            \midrule
            8 & 3.71 & 1.04 & 6.04 & 1.77\\
            9 & 3.28 & 0.879 & 6.43& 1.36 \\
            10 & 3.67 & 0.920 & 6.39 & 1.21 \\
            11 & 11.48 & 1.15 & 11.97 & 1.31 \\
            12 & 5.33 & 0.867 & 7.19 & 0.928 \\
        \bottomrule
    \end{tabular}
    \caption{Performance on the early week of a product cycle}
    \label{tab:start}
\end{table}

\section{Conclusion}
\label{sec:Conclusion}
Improving demand forecasting in E-commerce is possible through the use of global methods, which shares information between times series. In our paper, we proposed to use a gradient boosting method to do so. It allows us to exploits cross-features and non-linear effects that exists in the E-commerce data. Moreover, it also us to performs \emph{cold-start} prediction, with very few history on our products. 

We also proposed several tricks to tackle the difficulties inherent to E-commerce data. In particular, we proposed a way to compute seasonality for product thanks to the behavior of the rests of our products.

Finally, we evaluate our methodology on a real-world data-set, with a realistic number of products and we outperforms state of the art solutions for demand forecasting.

\bibliographystyle{apalike}
\setlength{\itemindent}{-\leftmargin}
\makeatletter\renewcommand{\@biblabel}[1]{}\makeatother
\bibliography{cap2017}

\end{document}